\documentclass{article}
\usepackage{ijcai24}
\usepackage{colortbl}
\usepackage{times}
\usepackage{soul}
\usepackage{url}
\usepackage[hidelinks]{hyperref}
\usepackage[utf8]{inputenc}
\usepackage[small]{caption}
\usepackage{graphicx}
\usepackage{amsmath}
\usepackage{amsthm}
\usepackage{booktabs}
\usepackage{algorithm}
\usepackage{algorithmic}
\usepackage[switch]{lineno}
\usepackage{hyperref}
\usepackage{tabularx, booktabs}
\usepackage{tikz}
\usepackage{arydshln}
\usepackage{CJKutf8}
\usepackage{subfig}
\usepackage{multirow}
\usepackage{enumitem}
\usepackage{makecell}
\usepackage{threeparttable}
\usepackage{amsfonts,mathtools}
\usepackage{mathrsfs}
\usepackage{float}

\title{Kuaiji: the First Chinese Accounting Large Language Model}

\author{
Jiayuan Luo$^1$\and
Songhua Yang$^2$ \and
Xiaoling Qiu$^1$\and
Panyu Chen$^3$ \and
Yufei Nai$^{4,}$$^6$\and \\
Wenxuan Zeng$^5$\and 
Wentao Zhang$^6$\footnote{Wentao Zhang is the corresponding author.}\and
Xinke Jiang$^5$\\
\affiliations
$^1$BAIC GROPU, Beiqi Foton Motor Co., Ltd.,
$^2$Wuhan University,
$^3$Hongkong University, \\
$^4$Jiangnan University
$^5$Peking University,
$^6$Wuxi Wisdom Shenshi Technology
Co., Ltd\\
\emails
\{joyingluo, suprit, alcorqiu, panyuchen, yfnai\}@foxmail.com,
zwx.andy@outlook.com\\
\{wentaozhtt,
thinkerjiang\}@foxmail.com
}
\begin{document}

\maketitle

\begin{abstract}
Large Language Models (LLMs) like ChatGPT and GPT-4 have demonstrated impressive proficiency in comprehending and generating natural language. However, they encounter difficulties when tasked with adapting to specialized domains such as accounting. To address this challenge, we introduce Kuaiji, a tailored Accounting Large Language Model. Kuaiji is meticulously fine-tuned using the Baichuan framework, which encompasses continuous pre-training and supervised fine-tuning processes. Supported by CAtAcctQA, a dataset containing 15,677 genuine accountant-client dialogues, Kuaiji exhibits exceptional accuracy and response speed. Our contributions encompass the creation of the first Chinese accounting dataset, the establishment of Kuaiji as a leading open-source Chinese accounting LLM, and the validation of its efficacy through real-world accounting scenarios. 

\end{abstract}



\section{Introduction}
\label{introduction}

\textbf{Large Language Models (LLMs)}, such as ChatGPT~\cite{ChatGPT} and GPT-4~\cite{OpenAI2023GPT4TR}, have made remarkable strides in pivotal areas. Through extensive pre-training on massive text corpora, they've demonstrated exceptional performance across various downstream tasks~\cite{kaplan2020scaling}, highlighting their vast potential in understanding and generating natural language~\cite{vu2024gptvoicetasker}. However, despite their progress, LLMs face significant challenges, including the need to adapt to specific tasks and domains~\cite{ziegler2020finetuning,wang2023selfinstruct}, the risk of factual inaccuracies (hallucinations) \cite{he2022rethinking}, and limitations in handling highly specialized queries~\cite{kandpal2023large}. These challenges compromise their reliability, especially in fields where accountability and trustworthiness are paramount, such as biomedicine~\cite{ji2023survey,song2024typing}, law~\cite{chatlaw1,chatlaw2}, and finance~\cite{zhang2023instructfingpt,yang2023fingpt}. Nonetheless, there remains a gap in LLMs tailored for accounting professionals. While many major companies are working on this, most of these models remain closed-source, resulting in a lack of publicly available specialized models for the accounting domain and limiting the widespread application of LLMs in this field.

As in accounting field, people use accounting to track their financial behavior and help them with financial decisions~\cite{penman2010accounting}. It’s common sense that accounting information is used by investors and managers to track the firm’s performance and value the company. Accounting information disclosure could change the stock price rapidly~\cite{zhu2016investor}. However, iterations of accounting regulations and vague accounting books have become a problem for people to use accounting information, and understand the relationship of this information with stock price~\cite{sunder1973relationship}. In addition, there is a gap between accounting research and practice, although university education can partly narrow the existing gap, a better solution is needed to bridge the gap~\cite{almaan2015difficulties}.

In this paper, we endeavor to address the challenges previously mentioned by explicitly fine-tuning a specialized Accounting Large Language Model, named \textbf{Kuaiji}. Based on the Baichuan framework, Kuaiji covers the entire pipeline from continuous pre-training to supervised fine-tuning. 
Accompanying this, we curated CAtAcctQA, a dataset comprising 70,000 QA pairs from real accountant-client dialogues across 14 accounting departments. Through three stages of model construction, including continuous pre-training, adaptive learning with domain expert feedback, and optimization, Kuaiji achieved remarkable performance. Evaluated against GPT-4 and human experts, Kuaiji excelled in accuracy and response speed, outperforming other open-source Chinese accounting language models. 
In summary, our contributions are listed as follows:
\begin{itemize} [leftmargin=*]
    \item To the best of our knowledge, we have created the inaugural Chinese accounting dataset, featuring 15,677 real instances spanning 14 distinct accounting subareas.
    \item Kuaiji stands as the premier open-source Chinese accounting LLM designed explicitly for accounting tasks and serves as a huge knowledge base and inference tools.
    \item Our validation of Kuaiji, conducted through real-world accounting cases, showcases its advantages, as suggested by accounting professionals.
\end{itemize}

\section{Related Work}
\label{related work}
\textbf{Large Language Models. }
The advent of Large Language Models (LLMs), exemplified by breakthroughs like ChatGPT~\cite{ChatGPT} and GPT-4~\cite{OpenAI2023GPT4TR}, has ignited a transformative wave in artificial intelligence research. These models have not only revolutionized natural language understanding and generation but have also paved the way for groundbreaking advancements across diverse applications. Despite the proprietary nature of some LLMs' training methodologies, the emergence of open-source alternatives such as Baichuan~\cite{baichuan}, LLaMA~\cite{touvron2023llama}, Bloom~\cite{scao2022bloom,workshop2023bloom}, and Falcon~\cite{penedo2023refinedweb} has democratized access to state-of-the-art language modeling capabilities.
Generative LLMs, or Language Models, exhibit remarkable capabilities in generating text that is both coherent and contextually relevant. Through pretraining on extensive text corpora and fine-tuning for alignment with human instructions, they excel in producing human-like text in response to given prompts or inputs.
In essence, LLMs operate by modeling the probability of a sentence $s$ (comprising a sequence of word tokens $q_1, q_2, \ldots, q_n$) as $p(s) = \prod_{i}^n p(q_i \mid q_{<i})$. Here, $q_i$ represents the $i$-th token of the sentence $s$, while $q_{<i}$ denotes the partial word token sequence preceding the $i$-th step.

\textbf{Finetune. }The most direct approach to injecting domain-specific knowledge into LLM is to finetune these models using high-quality datasets, enabling them to possess domain-specific QA capabilities through supervised learning. HuatuoGPT~\cite{zhang2023huatuogpt} leverages a mixture of medical knowledge generated by ChatGPT and real-world medical consultation data to achieve performance surpassing that of ChatGPT in the field of Chinese medical knowledge QA. ChatDoctor~\cite{li2023chatdoctor} finetunes LLaMA with 100,000 real-world data samples from online medical consultation platforms and employs a web engine search to assist in generating responses. KALA~\cite{kang2022kala} incorporates domain-specific knowledge in the form of triplets from KGss into the finetuning process, mitigating training expenses by freezing certain parameters. It is worth noting, however, that the acquisition and balanced curation of high-quality datasets during the finetuning process can be quite challenging, which directly impacts the performance of the finetuned LLMs~\cite{zhang2023huatuogpt}. Besides, substantial computational resources are required for fine-tuning base models, presenting challenges for frequent knowledge updates.

\section{Dataset Construction}
\label{Dataset Construction}
In this section, we will introduce the datasets used for training the Kuaiji model. To enhance the model's training effectiveness, we constructed a Continuous Pre-training (CPT) dataset and a Supervised Fine-Tuning (SFT) dataset tailored for different training stages. For the construction of the CPT dataset, we carefully considered both the professionalism and breadth of the accounting domain. Through an in-depth search for professional-oriented data and extensive acquisition of related domain data, we curated a CPT dataset totaling 828MB in size (cf. Section ~\ref{c:cpd}). Similarly, in preparing the SFT dataset, we reorganized existing data to meet our dataset requirements and leveraged the assistance of existing strong Language Models (LLMs) to handle labor-intensive divergent data preparation tasks. This process culminated in a dataset structured into instructions, inputs, and outputs, containing 35,784 fine-tuning data entries  (cf. Section ~\ref{c:sft}).

\begin{figure}[ht]
  \centering
\includegraphics[scale=0.35]{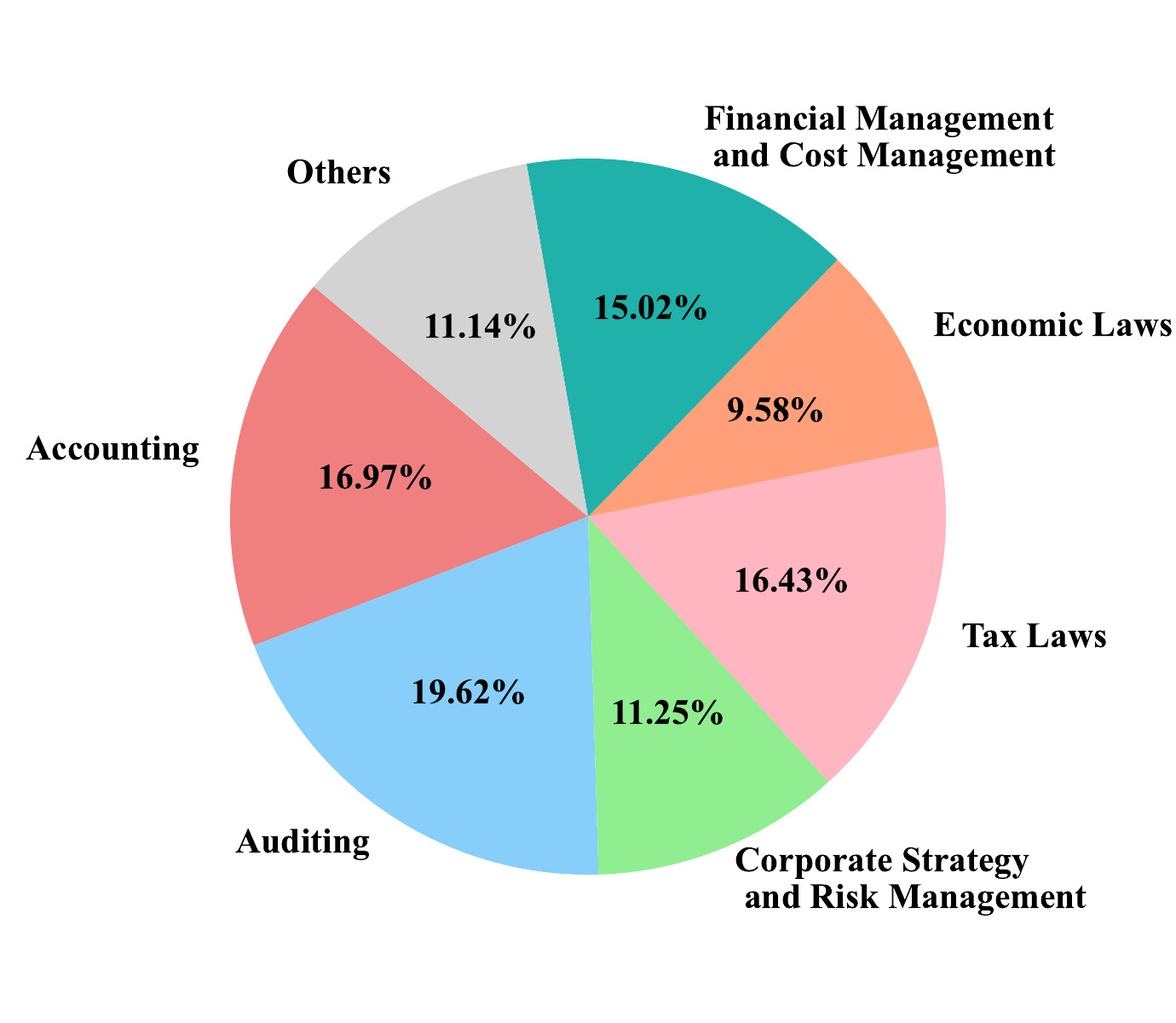}
  \caption{Statistics on the distribution of Kuaiji training dataset.}
  \label{fig:inrto_decline.png}
\end{figure}

We've organized our dataset following the criteria of the Chinese Certified Public Accountants (CPA) into six main categories: Accounting, Auditing, Corporate Strategy and Risk Management, Tax Laws, Economic Laws, and Financial Management and Cost Management. Additionally, there's a miscellaneous category labeled Others in Figure~\ref{fig:inrto_decline.png}.

\begin{figure*}[htb]
  \centering
\includegraphics[scale=0.41]{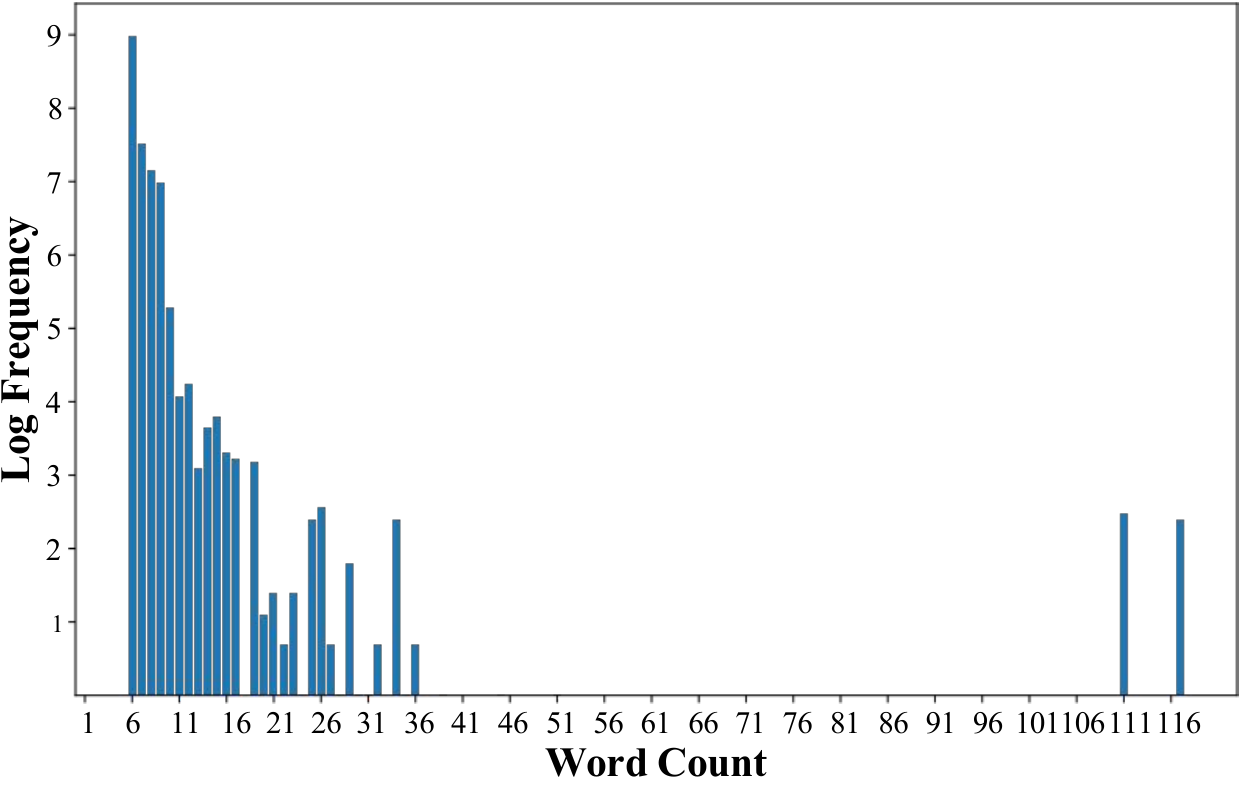} 
\includegraphics[scale=0.41]{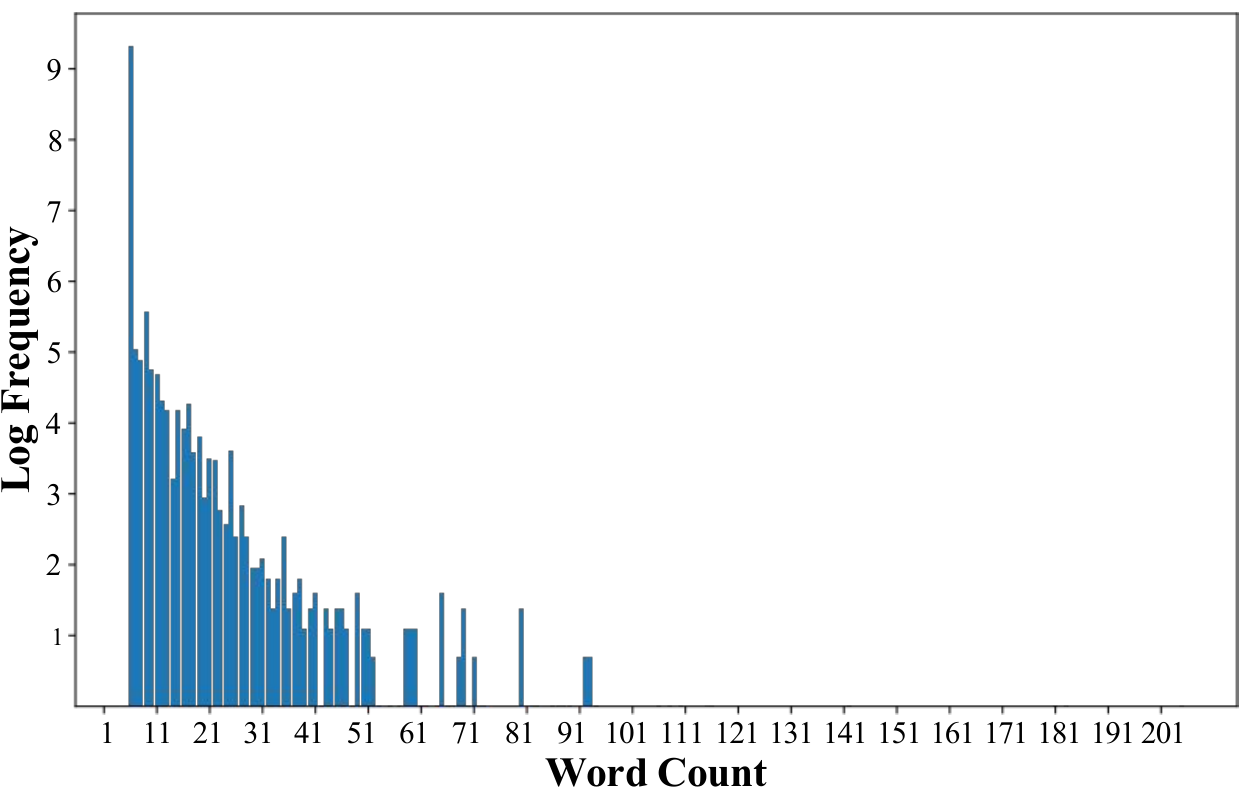}
  \caption{(\textbf{Left}.) SFT Dataset word count and log$_{10}$ frequency (Query). (\textbf{Right}.) SFT Dataset word count and log$_{10}$ frequency (Answer).
  }
   \label{fig:sft cal}
\end{figure*}

\subsection{Construction of Continuous Pre-training Dataset}
\label{c:cpd}
The accounting domain presents a demanding landscape characterized by its need for specialized expertise and its expansive scope covering diverse topics. Consequently, the construction of datasets for model development assumes paramount importance, necessitating a meticulous curation process. In our endeavor to meet the training requisites of our model, we placed particular emphasis on sourcing materials from authoritative sources within academia. Specifically, we selected university textbooks and academic papers from disciplines such as accounting, auditing, tax law, and economic law to form the foundation of our Continuous Pre-training dataset. Additionally, we incorporated supplementary materials from economics, business management, and corporate strategy to enrich the dataset further and ensure its professional integrity.

Diversification of the dataset was achieved through comprehensive data collection efforts spanning various channels. These channels included specialized financial investigation reports\footnote{https://www.ctrchina.cn/}, annual reports of publicly listed companies\footnote{https://www.cninfo.com.cn/}, national statistical bureaus \footnote{https://www.stats.gov.cn/}, financial commentaries from columns\footnote{https://finance.eastmoney.com/}, media coverage of financial news \footnote{https://finance.sina.com.cn/}, and other open datasets \footnote{https://www.chinairn.com/}. The diverse array of data sources not only enhances the dataset's breadth but also imbues it with real-time relevance, mirroring the complexities of real-world scenarios. Moreover, these sources serve as complementary resources to traditional academic materials, fostering a comprehensive understanding of professional knowledge within the model.

In tandem with efforts to enhance dataset diversity, rigorous data cleansing procedures were implemented to ensure the dataset's quality and integrity. This involved meticulous error correction, noise reduction, and consistency checks. Additionally, stringent measures were taken to safeguard data privacy, including the removal of sensitive information and anonymization techniques. Furthermore, expert validation from professionals spanning accounting, economics, law, and related fields was sought to provide an additional layer of scrutiny, thereby upholding the dataset's professionalism and accuracy. 

In the end, we divided the obtained data into two categories: specialized and related domains, and mixed them in a ratio of 7:3. This mixture helps to expand the training dataset, accelerate the convergence speed of the model, and enhance its robustness. The types and sizes of the datasets are presented in  Table~\ref{tab:CPT_dataset}.

\begin{table}[t!]
\small
\centering

\renewcommand{\arraystretch}{1.4}
\begin{tabularx}{\linewidth}{>{\centering\arraybackslash}Xlll}
  \toprule
  \textbf{Dataset} & \textbf{Type} & \textbf{Size} \\
  \midrule
  University Textbook & Specialized & 324MB \\
  Academic Paper & Specialized & 206MB \\
  Accounting Wiki & Wiki Data  & 78MB \\
  Company Financial Statements & Related field & 173MB \\
  National Statistical Data & Related field & 7MB \\
  Financial Survey Report & Related field & 16MB \\
  Financial News and Commentary & Related field & 24MB \\
  \bottomrule
\end{tabularx}
\caption{Kuaiji pre-training data statistics and sources.}
\label{tab:CPT_dataset}
\end{table}


\subsection{Construction of Supervised Fine-Tuning Dataset}
\label{c:sft}
Fine-tuning is crucial for the training of large models, as it enables models to better understand and generate text relevant to specific domains or tasks by training on domain-specific data. Due to the scarcity of fine-tuning datasets in certain domains, we opted to utilize domain-specific practice exercises and exam papers to aid in dataset construction (cf. Section~\ref{sec:from text}). 
However, because exam questions mainly focus on concept and scenario questions, there is a lack of understanding in more fields. Although the quality of the dataset constructed through exam questions is excellent, collecting exam questions is extremely labor-intensive and the amount of data is limited.
To address this shortfall, we devised a specialized prompt following Self-Instruct Prompt~\cite{selfinstruct} to assist in generating the fine-tuning dataset (cf. Section~\ref{sec:from llms}) with the help of strong LLMs given reference domain text, i.e. GPT. The constructed instructions can be referred to in Table~\ref{tab:Fine-tuning Data Example}.

%

\subsubsection{From Test}
\label{sec:from text}
In the selection of our original dataset, we included practice questions from various sources, including the Certified Practising Accountant (CPA) exam for junior, intermediate, and senior levels, graduate entrance exams for accounting and management fields, as well as exercise collections from professional textbooks and industry-standard problem compilations. To ensure the accuracy of dataset instructions and input-output pairs, we excluded ambiguous questions that might lead to ambiguous outputs during the dataset construction process. Instead, we focused on selecting questions with clear and definitive answers. Moreover, in our design process, we aimed to cover various aspects and difficulty levels relevant to the model's learning requirements and target tasks comprehensively.

Similarly, expert validation was conducted to ensure the clarity and conciseness of the questions, minimizing ambiguity and ensuring the explicit understandability of the tested content. We prepared a total of 3,497 instructions derived from these practice question collections.

\subsubsection{From Strong LLMs}
\label{sec:from llms}
We selected our self-constructed Continuous Pre-training dataset and supplemented it with fine-tuning data generated by Strong LLMs, i.e. GPT, as they are trained under huge and various general knowledge, such as high-quality literature corpus and dialogues, etc., which is beneficial for comprehending the accounting knowledge and generating new dialogues without domain-specific knowledge.

Initially, we crafted the self-instruction prompt, as illustrated in Table~\ref{tab:self-instruction}, to automate the generation of instructions, inputs, and outputs. This process involves analyzing the input domain text and formatting the output accordingly, using the domain-specific text dataset as a reference. To facilitate efficient text analysis, we introduced a sliding window mechanism. This mechanism allows GPT to comprehend the domain text within a fixed window size—measured in lines—and slide over the text by a predetermined distance to continue reading. It is important to highlight that we directed the LLMs to strictly follow given Requirements and Reference Formats, aiming to produce instructions that are not only diverse but also detailed.

This methodology ensures the relevance of the generated content to the context, significantly improving logical coherence, consistency, and accuracy, while also reducing the likelihood of semantic discontinuities. Through comparative analyses, we determined the ideal window size and sliding stride to be 50 lines and 30 lines, respectively. This optimization enhances the effectiveness of our approach in generating contextually relevant and coherent content.

\begin{table*}[!h]
    \centering
    \caption{Self-Instruction Prompt.}
    \label{tab:self-instruction}
    \begin{tabularx}{\textwidth}{X}
        \toprule
        \textbf{[Task Description]} 
        
        You are currently an expert in the field of accounting and finance and have been tasked with providing \textbf{10} diverse task instructions based on [Background Knowledge]. The following [Requirements] outline the criteria you need to meet when providing instructions, as detailed in the [Reference Format].
        
        \\
    \textbf{[Requirements]} 
        
        \underline{1)} Avoid repeating verbs in each instruction and strive for diversity in both instructions and tone. 
        
        \underline{2)} Instructions should encompass various types of tasks, including brainstorming, open and closed QA, rewriting, extraction, generation, classification, chat, and summarization. 
        
        \underline{3)} Instructions should be in Chinese and consist of 1 to 2 sentences. Imperative or interrogative sentences are allowed, without line breaks.
        
        \underline{4)} You should generate appropriate input for the instruction, including specific examples that involve real data to make the instruction challenging. 
        
        \underline{5)} Not all instructions require input. For example, for common sense information inquiries, such as ``what is the highest mountain in the world," simply state ``No Input" in the input field. Text materials (e.g., articles, links) should provide examples directly in the input section. Other media types, like audio, images, videos, or links, do not meet the requirements.
        
        \underline{6)} The output should be an appropriate response to the instruction and input.
        
        \\
        \textbf{[Reference Format]}
        
[{
``Instruction": ``According to the provided accounting rules, classify the following words into the correct categories: Assets, Equity, Expenses, Revenue.",
``Input": ``Understanding the accounting rules: Debit increases and credit decreases assets and equity oppositely. Fixed assets are the same overall, remember details clearly and don't confused. Distinguish between profit and loss accounts; expenses and revenue are not the same.",
``Output": ``Assets: Fixed assets Equity: Debit increases and credit decreases Expenses: Expenses Revenue: Revenue"
},

\{
``Instruction": ``Based on the following text, summarize the important characteristics or definitions of the relationship between tax law and tax legislation.",
``Input": ``Main question: Is national taxation based on political power or property rights? Subquestion 1: Expression of tax law and tax legislation relationship.",
``Output": ``Tax law fundamentally embodies a distribution relationship. The content of tax legislation relationship is the most substantial aspect of tax legislation and the essence of tax law."
\}]

        \\
        
        \textbf{[Background Knowledge]} 
        
        Financial statement auditing is the core business of certified public accountants. A financial statement audit refers to the provision of reasonable assurance by a certified public accountant on whether the financial statements are free from material misstatement, actively expressing opinions, and enhancing the level of trust of expected users other than management in the financial statements...
        \\
        \bottomrule
    \end{tabularx}
\end{table*}

\begin{table*}[!h]
    \centering
    \caption{Fine-tuning Data Example.}
    \label{tab:Fine-tuning Data Example}
    \begin{tabularx}{\textwidth}{X}
        \toprule
        \textbf{[Fine-tuning Data Example]}

\{
``\textbf{Instruction}": ``List the ways in which investment properties can be converted and the corresponding accounting treatment, and briefly explain the principles of conversion.",
\\
\\

``\textbf{Input}": ``Method of conversion of investment properties, Accounting treatment, Explanation of conversion principles",
\\
\\

``\textbf{Output}": ``An enterprise that converts investment property means that the enterprise has conclusive evidence of a change in the use of its real estate, and should convert investment property to other assets if one of the following conditions is met, for example, if the investment property begins to be used for its own purposes. In the cost model, the book value of the real estate before conversion should be used as the recorded value after conversion. When an investment property measured by the fair value model is converted to real estate for its own use, the fair value on the date of conversion shall be used as the book value of the real estate for its own use. The difference between the fair value and the original book value shall be recognized in profit or loss for the current period.
Under the cost model, in case of conversion of investment property to owner-occupied property, the corresponding accounting treatment is: \\
Debit: Fixed assets (intangible assets)\\
        Accumulated depreciation (accumulated amortization) of investment properties\\
        Provision for impairment of investment properties\\
Credit: Investment properties - cost\\
        Accumulated depreciation (accumulated amortization)\\
        Provision for impairment of fixed assets (intangible assets)\\
The conversion principles are as follows: \\
1. Comparability principle: The conversion form and conversion date should be consistent with previous accounting periods to ensure the comparability of financial statements.

2. Reliability principle: The conversion form and date should be based on reliable evidence and data to ensure the accuracy and reliability of the conversion process.

3. Consistency principle: Once a certain conversion form and conversion date are selected, it should be consistent in future accounting periods unless there are sufficient reasons for change.

4. Dominance principle: The selection of the conversion form and conversion date should be based on the dominant characteristics and economic substance of investment real estate to reflect its true value and financial condition.

5. Economic practicability principle: The selection of the conversion form and conversion date should take into account economic practicability to meet users' information needs and decision-making purposes."
\}
\\
\bottomrule
    \end{tabularx}
\end{table*}

\section{Methodology}
\label{method}
\begin{figure*}[t!]
\centering
\includegraphics[width=\textwidth]{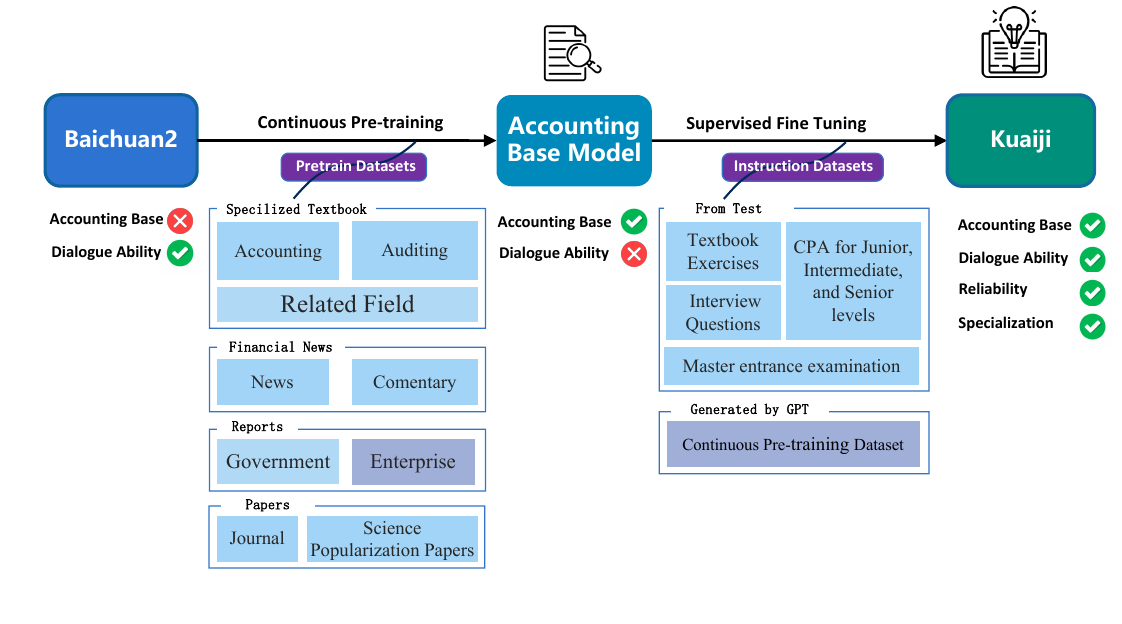}
    \caption{The overall flowchart of constructing Kuaiji.}
    \label{fig:main_model.png}
\end{figure*}

This section delves into the fine-tuning process of Kuaiji, spanning two key stages: Continuous Pre-Training (CPT) in Section~\ref{sec: cpt} and Supervised Fine-Tuning (SFT) in Section~\ref{sec: sft}. Each stage is discussed sequentially to mirror the research workflow. Additionally, we provide a comprehensive method flowchart, depicted in Figure~\ref{fig:main_model.png}, to offer a visual representation of the methodology employed in Kuaiji. The whole training framework adopts QLoRA strategy as stated in Section~\ref{sec: qlora}.

\subsection{Continuous Pre-training}
\label{sec: cpt}
The high-quality pre-training corpus can greatly improve the performance of LLM and even break the scaling laws to some extent \cite{gunasekar2023textbooks}. Among them, continuous pre-training is a crucial phase where the Kuaiji model undergoes extensive training on vast and diverse unlabeled datasets. This process spans multiple iterations, each aimed at refining the model's language understanding capabilities. Initially, the Kuaiji is initialized with the pre-trained weights of basic LLMs and learns to predict missing words or segments within sentences using self-supervised learning objectives such as masked language modeling (MLM) and next-sentence prediction (NSP). Through exposure to a wide variety of textual data sources, as suggested in Section~\ref{sec: cpt}, the model gradually acquires a rich domain understanding of language structure, semantics, and context in the accounting domain. Additionally, techniques like attention mechanisms and multi-layer architectures are employed to capture complex linguistic patterns and dependencies. 

\subsection{Supervised Fine-Tuning}
\label{sec: sft}
Supervised fine-tuning is a pivotal step in refining the Kuaiji model for specific accounting tasks. Leveraging task-specific labeled datasets from both human and strong LLMs, such as the meticulously curated CAtAcctQA dataset and the distilled data from GPT, this phase focuses on adapting the pre-trained model to the intricacies of accounting language and domain-specific conventions. With high-quality domain dialogue data, the model can effectively invoke the accounting knowledge accumulated during pre-training, thereby understanding and responding to users' queries.
SFT involves adjusting the model's parameters through gradient-based optimization techniques, guided by the labeled data's ground truth annotations. By iteratively feeding batches of labeled examples to the model and updating its weights through back-propagation, the model learns to refine its internal representations to better align with the target accounting tasks. This process allows Kuaiji to capture nuances in accounting terminology, understand contextual cues unique to financial documents, and effectively handle domain-specific queries and tasks. The supervised fine-tuning phase is instrumental in enhancing Kuaiji's accuracy, responsiveness, and overall performance in accounting language processing, making it well-suited for real-world applications in finance and accounting domains.
Moreover, to avoid the collapse of inherent capabilities rather than learning substantive ones \cite{gudibande2023false,shumailov2023curse}, we mix the domain-specific instructions with general ones, with the proportion of 80\%:20\%~\cite{chen2023skillit,dong2024abilities}.

We have conducted statistics on the token count and logarithmic frequency of the queries and answers of the instructions we generated, and the results are shown in Figure~\ref{fig:sft cal}.

\subsection{Training Strategy: QLoRA}
\label{sec: qlora}
During model training, we adopt QLoRA due to its ability to mitigate memory limitations and performance degradation encountered during CPT and SFT phases. By leveraging low-precision quantization and optimized memory management, QLoRA effectively minimizes memory usage while preserving model accuracy. This empowers us to efficiently fine-tune models even on hardware-constrained devices, facilitating the deployment and utilization of large-scale models in practical applications.


In the modified version of QLoRA that incorporates 8-bit NormalFloat quantization, based on Quantile Quantization and Block-wise Quantization, the parameters of the neural network are mapped to an 8-bit representation. This enhances the precision by offering 256 possible values for each parameter, compared to the 16 values provided by the original 4-bit representation. The main goal remains to preserve the distribution of the original parameters by dividing the parameter space into quantiles, each mapped to a specific 8-bit value. The definition of QLoRA for a single linear layer in the quantized base model with a single LoRA adapter, adjusted for 8-bit quantization, is as follows:
\begin{equation}
\begin{aligned}
Y^{\text{BF16}} = X^{\text{BF16}} \cdot \text{doubleDequant}(c^{\text{FP32}}, c^{k\text{-bit}}, W^{\text{NF8}}) +  \\ X^{\text{BF16}} \cdot L^{\text{BF16}}_1 \cdot L^{\text{BF16}}_2
\end{aligned}
\end{equation}

Where $\text{doubleDequant}(\cdot)$ is adapted for 8-bit quantization as:
\begin{equation}
\begin{aligned}
\text{doubleDequant}(c^{\text{FP32}}, c^{k\text{-bit}}, W^{k\text{-bit}}) = \text{dequant} \bigl(\text{dequant} \\(c^{\text{FP32}}, c^{k\text{-bit}}), W^{\text{8bit}}\bigl) = W^{\text{BF16}}.
\end{aligned}
\end{equation}
In adopting 8-bit quantization (NF8) for weight parameters $W$, we aim to strike an optimal balance between enhanced precision and efficient memory use, with the possibility of adjusting $W$ and $c_2$ block sizes tailored to specific application needs and computational constraints. Priority is given to updating adapter weight gradients $\frac{\partial E}{\partial L_i}$, thus sidestepping the computation of gradients for the 8-bit quantized weights $\frac{\partial E}{\partial W}$. The process involves de-quantizing weights from their NF8 storage representation to the BF16 computational format to compute the derivative $\frac{\partial X}{\partial W}$ with BFloat16 precision, ensuring precision in gradient calculation without excessive memory demand.


\section{Empirical Analysis}
\label{experiments}
\subsection{Training Details}
Our model is based on Baichuan2-13B-chat, a specialized Chinese Large Language Model (LLM) consisting of 13 billion parameters, specifically designed for conversational tasks. The training process utilizes 4 A100-80G GPUs with parallelization techniques within the Baichuan2 framework. 
During training, we implement the low-rank adaptation (LORA) parameter-efficient tuning method with a rank equal to 8 to optimize model performance. This methodology is facilitated through the Transformers4 and PEFT5 libraries. To ensure efficient resource utilization, we adopt fp16 precision with ZeRO-2, along with gradient accumulation strategies. Additionally, to manage computational complexity, we limit the length of individual responses to 4096 tokens. Optimization techniques include the AdamW optimizer, a dropout rate of 0.1, and a cosine learning rate scheduler. A portion of the training dataset (10\%) is reserved for validation, and the best-performing checkpoints are selected as the final model. Mechanisms are also in place to handle loss during gradient explosions and to adjust the learning rate for training stability. The hyperparameters for each stage of training are iteratively refined to ensure convergence within an effective loss range. 

During the model training process, we set the learning rate to 5e-05 and utilized a cosine annealing scheduler to adjust the learning rate dynamically. Regarding gradient clipping, we applied a threshold of 1.0 to prevent gradient explosion and enhance training stability. To control resource wastage during training, we constrained the maximum input length to 512 tokens. This approach is feasible because the majority of information is typically contained within the initial few tokens of a sequence. Limiting the maximum length ensures that the model focuses more on crucial information, thereby mitigating overfitting and unnecessary computations. 

\subsection{Baselines}
To conduct a thorough evaluation of our model, we have chosen a variety of LLMs with different parameter scales as baselines for comparison. These include both general-purpose and account-specific LLMs:
\begin{itemize} [leftmargin=*]
\item \textbf{ChatGPT}~\cite{ChatGPT}: This widely recognized LLM boasts approximately 175 billion parameters. While not explicitly trained for medical applications, it has demonstrated remarkable performance in account dialogue tasks. 
\item \textbf{Baichuan2-13B}~\cite{baichuan}: We also include the base Baichuan2-13B for comparison.
\end{itemize}

\subsection{Case Study}
To demonstrate the effectiveness of accounting, we compared Kuaiji and two other LLMs in three case studies.

\begin{table*}[!h]
    \centering
    \caption{Case study 1.}
    \label{tab:cases1}
    
    \begin{tabularx}{\textwidth}{p{0.2\textwidth}X}
        \toprule
        \textbf{Query}  & 
        \textit{How are the three accounting statements linked in the valuation model?} \newline \\ \hline
        \textbf{ChatGPT} & \textit{
The income statement, balance sheet, and cash flow statement are interconnected in the valuation model through adjustments and transactions that flow between them, such as net income affecting retained earnings on the balance sheet and operating cash flows derived from net income on the cash flow statement.}
        \newline \\ \hdashline
        \textbf{Baichuan2-13B} &
        \textit{The three accounting statements linked in the valuation model are the income statement, the balance sheet, and the cash flow statement.} 
        \newline \\ \hdashline
    \textbf{Kuaiji} & \textit{In the valuation model, there is a close relationship between the three accounting statements (i.e. balance sheet, income statement, and cash flow statement). Firstly, the balance sheet reflects the assets, liabilities, and owner's equity of a company, reflecting its financial condition and debt paying ability. The income statement reflects the company's revenue and expenses, including its profit and loss situation. The cash flow statement reflects the company's cash inflows and outflows, including cash flows from investments, financing, and operating activities. The connection between these three statements is that they all reflect the financial condition and operating results of the company, but use different financial indicators and accounting methods. By integrating these three reports together, investors can have a comprehensive understanding of the company's financial condition and profitability, enabling more accurate valuation analysis.}
        \newline \\
        \bottomrule
    \end{tabularx}
    \label{tab:case study1}
\end{table*}

\begin{table*}[!h]
    \centering
    \caption{Case study 2.}
    \label{tab:cases2}
    
    \begin{tabularx}{\textwidth}{p{0.2\textwidth}X}
        \toprule
        \textbf{Query}  & 
        \textit{Assuming a tax rate of 25\%, how do the three statements change when one hundred dollars of new depreciation is added?} 
        \newline \\ \hline
        \textbf{ChatGPT} & \textit{
Adding \$100 of new depreciation would decrease taxable income by \$100, resulting in a tax savings of \$25 (25\% of \$100).}
        \newline \\ \hdashline
        \textbf{Baichuan2-13B} &
        \textit{Adding \$100 of new depreciation will result in a reduction in pre-tax income, a higher depreciation asset on the balance sheet, and a decrease in operating cash flow, offset by an increase in investing cash flow.
} 
        \newline \\ \hdashline
    \textbf{Kuaiji} & \textit{When a new depreciation of \$100 is added, all three reports will undergo changes. Firstly, in the balance sheet, the book value of fixed assets will decrease by \$100, and accumulated depreciation will increase by \$100. Secondly, in the income statement, depreciation expenses will increase by \$100, and the impact on profit depends on whether depreciation expenses belong to expensed depreciation. If the depreciation expense belongs to expensed depreciation, the total profit will increase by \$100; If depreciation expenses belong to capitalized depreciation, the total profit will not change significantly. Finally, in the cash flow statement, depreciation expenses will increase by \$100, affecting the cash flows generated from operating activities. Therefore, an increase of \$100 in new depreciation will affect the company's financial condition, operating results, and cash flow, and the impact on the financial statements needs to be comprehensively considered.}
        \newline \\
        \bottomrule
    \end{tabularx}
    \label{tab:case study2}
\end{table*}

\begin{table*}[!h]
    \centering
    \caption{Case study 3.}
    \label{tab:cases3}
    
    \begin{tabularx}{\textwidth}{p{0.2\textwidth}X}
        \toprule
        \textbf{Query}  & 
        \textit{If you could only use two accounting statements to analyze a company, which two would you choose? Why?}
        \newline \\ \hline
        \textbf{ChatGPT} & \textit{
I would choose the income statement and the cash flow statement because they provide essential insights into a company's profitability and cash flow dynamics, offering a comprehensive view of its financial performance and sustainability.
}
        \newline \\ \hdashline
        \textbf{Baichuan2-13B} &
        \textit{If I could only use two accounting statements to analyze a company, I would choose the income statement and the balance sheet. Both statements provide valuable information about the company's financial health and performance.
} 
        \newline \\ \hdashline
    \textbf{Kuaiji} & \textit{If you can only use two financial statements to analyze a company, I would choose the balance sheet and cash flow statement. The reasons are as follows:
1. The balance sheet can reflect the financial condition of a company, including assets, liabilities, and owner's equity, which helps us understand the company's asset structure and financial strength. Through the balance sheet, we can calculate financial indicators such as the company's debt-to-equity ratio and asset turnover to evaluate the company's debt-paying ability and operational efficiency.
The cash flow statement reflects the company's cash inflows and outflows, including cash flows from investments, financing, and operating activities. Through the cash flow statement, we can understand the company's cash sources and purposes, as well as the cash flow situation. This helps us evaluate the company's cash liquidity and cash flow from operating activities.
Therefore, the balance sheet and cash flow statement can provide us with the financial and cash flow situation of the company, and the combination of these two statements can help us comprehensively evaluate the financial and operating conditions of the company.}
        \newline \\
        \bottomrule
    \end{tabularx}
    \label{tab:case study3}
\end{table*}

\begin{figure*}[t]
\centering
\includegraphics[width=1\textwidth]{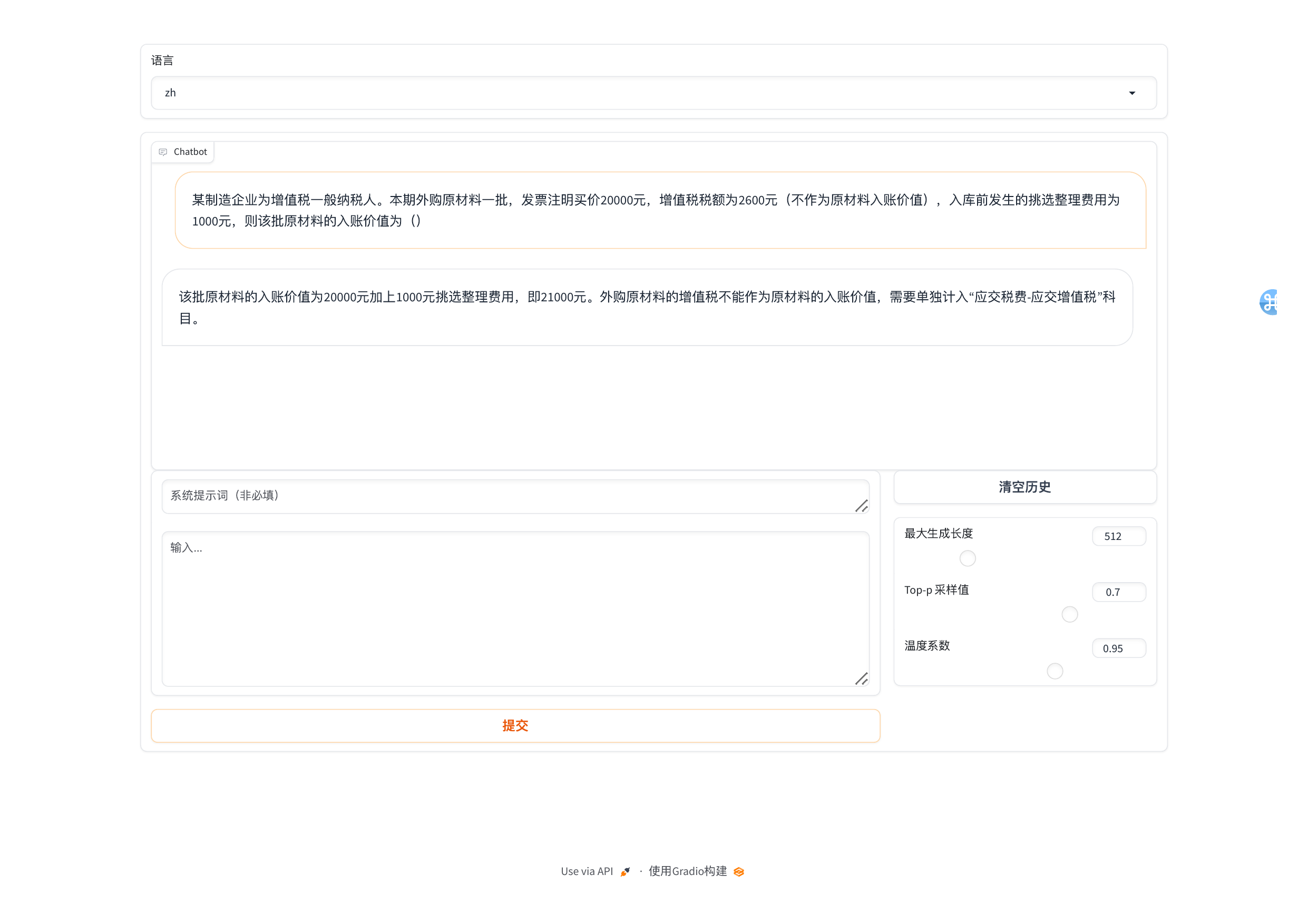}
    \caption{Kuaiji sigle-turn Q\&A test (in Chinese).}
\label{fig:kuaiji_1}
\end{figure*}

\begin{figure*}[t]
\centering
\includegraphics[width=1\textwidth]{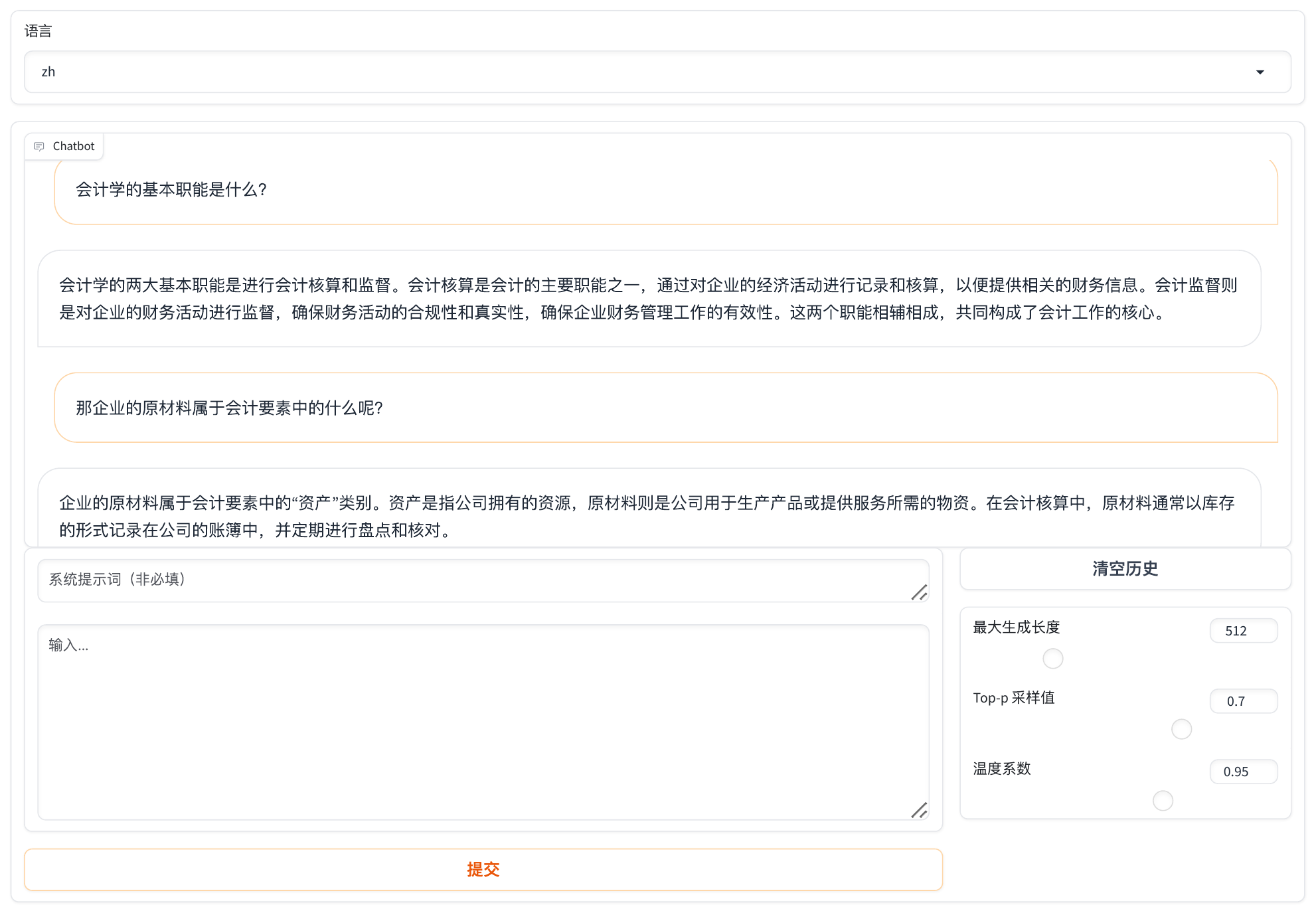}
    \caption{Kuaiji multi-turns Q\&A test (in Chinese).}
\label{fig:kuaiji_2}
\end{figure*}

\subsection{Application Scenarios}
After training on a large number of accounting professional datasets, our accounting model greatly assists in the following subfields:

\begin{enumerate}[label=\textbf{\arabic*.}]
    \item \textbf{Financial Analysis and Reporting:} Kuaiji can analyze company financial statements, provide interpretations of financial indicators, and identify potential risks and opportunities. This is crucial for businesses to understand their financial health, make informed decisions, and communicate effectively with stakeholders.
    
    \item \textbf{Tax Consultancy:} Kuaiji can offer guidance and advice on tax laws, assisting businesses in compliance and optimizing tax strategies. Given the complexity of tax regulations, having a reliable tool like Kuaiji can ensure accurate tax filings and minimize the risk of penalties or audits.
    
    \item \textbf{Risk Management:} Utilizing Kuaiji, companies can identify and assess financial risks, receive risk management recommendations, and reduce their exposure to risks. By proactively managing risks, businesses can safeguard their assets, reputation, and long-term viability.
    
    \item \textbf{Management Decision Support:} Kuaiji can analyze and interpret financial data, provide decision support to management, and aiding in strategic planning, and business implementation. With Kuaiji's insights, executives can make data-driven decisions, allocate resources effectively, and drive organizational growth.
    
    \item \textbf{Audit and Compliance:} Kuaiji can assist in auditing processes, identifying anomalies and potential non-compliance issues, and ensuring adherence to relevant regulations and standards. This helps companies maintain transparency, integrity, and trust among investors, regulators, and the public.
    
    \item \textbf{Education and Training:} Kuaiji can serve as an educational tool, offering learning resources and guidance for students in accounting and finance, aiding in the understanding of complex accounting principles and practices. By leveraging Kuaiji, educators can enhance the learning experience and equip future professionals with practical skills.
    
    \item \textbf{Market Forecasting and Investment Decision Making:} Kuaiji can analyze financial data and market trends to forecast future market movements, providing guidance for investment decisions. In a volatile market environment, having accurate predictions from Kuaiji can help investors mitigate risks and capitalize on opportunities.
    
    \item \textbf{Financial Planning and Consultancy:} Kuaiji can provide customized financial planning and advice based on individual or business financial situations, helping them achieve their financial goals. Whether it's retirement planning, debt management, or wealth accumulation, Kuaiji can offer personalized strategies for financial success.
\end{enumerate}

Our Kuaiji can achieve multiple rounds of dialogue and retain memory function.
We chose an accounting calculation question to test Kuaiji's response in Figure~\ref{fig:kuaiji_1}, and in addition, we also chose a multi-round dialogue task to demonstrate Kuaiji's effectiveness in Figure~\ref{fig:kuaiji_2}.

\section{Conclusion and Future Work}

In summary, although Large Language Models (LLMs) such as ChatGPT and GPT-4 have showcased remarkable capabilities in natural language tasks, their adaptation to specialized fields like accounting poses significant challenges. In response, we have introduced Kuaiji, a finely-tuned Accounting Large Language Model developed within the Baichuan framework. Through continuous pre-training and supervised fine-tuning, Kuaiji has exhibited outstanding accuracy and responsiveness, underpinned by the CAtAcctQA dataset. Our endeavors have led to the creation of the inaugural Chinese accounting dataset and the establishment of Kuaiji as a premier open-source Chinese accounting LLM, validated through real-world accounting scenarios.
Looking forward, our future endeavors will concentrate on integrating Reinforcement Learning from Human Feedback (RLHF) to enhance model refinement. Moreover, efforts will be dedicated to expanding the dataset to encompass a broader and more diverse range of data, ensuring Kuaiji's ongoing enhancement and effectiveness in processing accounting language across diverse applications.

\bibliographystyle{named}
\bibliography{ijcai24}

\end{document}